\documentclass{article}

\PassOptionsToPackage{numbers,sort,compress}{natbib}

\usepackage[preprint]{open_neurips}

\usepackage[utf8]{inputenc} 
\usepackage[T1]{fontenc}    
\usepackage{url}            
\usepackage{booktabs}       
\usepackage{microtype}      
\usepackage{listings}
\usepackage{overpic}
\usepackage{wrapfig}
\usepackage{graphicx}
\usepackage{colortbl}
\usepackage{makecell}
\usepackage{multirow}
\usepackage{tabularx}
\usepackage{verbatim}

\usepackage[pagebackref=true,breaklinks=true,colorlinks,bookmarks=false,citecolor=blue,linkcolor=blue]{hyperref}

\usepackage{amsthm}
\usepackage{amssymb}
\usepackage{amsmath}
\usepackage{amsfonts}       
\usepackage{nicefrac}       
\usepackage{mathtools}

\usepackage[capitalize]{cleveref}
\crefname{equation}{Eq.}{Eq.}
\crefname{figure}{Fig.}{Fig.}
\crefname{table}{Tab.}{Tab.~}
\crefname{section}{Sec.}{Sec.~}
\crefname{algorithm}{Alg.}{Alg.~}
\crefname{thm}{Theorem}{Theorem~}
\crefname{lemma}{Lemma}{Lemma~}
\crefname{appendix}{Appendix}{Appendix~}

\theoremstyle{plain}
\newtheorem{theorem}{Theorem}[section]

\theoremstyle{definition}

\theoremstyle{remark}

\usepackage[textsize=tiny]{todonotes}

\def\ie{\textit{i.e.,~}}
\def\eg{\textit{e.g.,~}}
\def\etc{\textit{etc}}

\def\wrt{\textit{w.r.t.~}}

\def\sota{state-of-the-art~}

\newcommand{\tbf}[1]{\textbf{#1}}
\newcommand{\ul}[1]{\underline{#1}}

\usepackage{amsmath,amsfonts,bm}

\def\eqref#1{equation~\ref{#1}}

\def\1{\bm{1}}

\def\rvb{{\mathbf{b}}}
\def\rvc{{\mathbf{c}}}

\def\rvf{{\mathbf{f}}}
\def\rvg{{\mathbf{g}}}
\def\rvh{{\mathbf{h}}}
\def\rvu{{\mathbf{i}}}

\def\rvk{{\mathbf{k}}}

\def\rvq{{\mathbf{q}}}

\def\rvt{{\mathbf{t}}}
\def\rvu{{\mathbf{u}}}
\def\rvv{{\mathbf{v}}}

\def\rvx{{\mathbf{x}}}
\def\rvy{{\mathbf{y}}}
\def\rvz{{\mathbf{z}}}

\def\rmA{{\mathbf{A}}}
\def\rmB{{\mathbf{B}}}
\def\rmC{{\mathbf{C}}}

\def\rmG{{\mathbf{G}}}

\def\rmI{{\mathbf{I}}}
\def\rmJ{{\mathbf{J}}}

\def\rmR{{\mathbf{R}}}

\def\rmU{{\mathbf{U}}}
\def\rmV{{\mathbf{V}}}
\def\rmW{{\mathbf{W}}}
\def\rmX{{\mathbf{X}}}

\def\rmZ{{\mathbf{Z}}}

\DeclareMathAlphabet{\mathsfit}{\encodingdefault}{\sfdefault}{m}{sl}
\SetMathAlphabet{\mathsfit}{bold}{\encodingdefault}{\sfdefault}{bx}{n}

\def\gB{{\mathcal{B}}}
\def\gC{{\mathcal{C}}}

\def\gL{{\mathcal{L}}}

\def\gN{{\mathcal{N}}}
\def\gO{{\mathcal{O}}}

\DeclareMathOperator*{\argmin}{arg\,min}

\def\L{\gL}
\def\f{f_\theta}
\def\gf{g_\theta}

\newcommand{\Rdim}[1]{\mathbb{R}^{#1}}

\def\zstar{\rvz^\star}
\def\z{\rvz}
\def\x{\rvx}
\def\y{\rvy}
\def\g{\rvg}

\def\Zstar{\rmZ^\star}
\def\Z{\rmZ}
\def\X{\rmX}
\def\W{\rmW}
\def\A{\rmA}
\def\B{\rmB}
\def\I{\rmI}

\def\c{\rvc}
\def\q{\rvq}

\def\u{\rvu}

\def\hstar{\rvh^\star}
\def\fstar{\rvf^\star}
\def\cstar{\rvc^\star}

\def\unet{\boldsymbol{\epsilon}_\theta}

\def\IFT{\frac{\partial \L}{\partial \theta}}
\def\PhantomGrad{\widehat{\frac{\partial \L}{\partial \theta}}}

\def\dLdz{\frac{\partial \L}{\,\partial \zstar}}
\def\dfdz{\frac{\partial \f}{\partial \zstar}}
\def\J{\left(I - \dfdz\right)}

\def\invJ{\J^{-1}}

\def\Jf{J_{\f}}

\title{TorchDEQ: A Library for Deep Equilibrium Models}

\author{%
    Zhengyang Geng\textsuperscript{1}
    \quad
    J. Zico Kolter\textsuperscript{1,2}
    \\\\
    $^1$CMU \quad $^2$Bosch Center for AI
}

\begin{document}

\maketitle

\begin{abstract}
Deep Equilibrium (DEQ) Models, an emerging class of implicit models that maps inputs to fixed points of neural networks, are of growing interest in the deep learning community.  
However, training and applying DEQ models is currently done in an ad-hoc fashion, with various techniques spread across the literature.   
In this work, we systematically revisit DEQs and present TorchDEQ, an out-of-the-box PyTorch-based library that allows users to define, train, and infer using DEQs over multiple domains with minimal code and best practices. 
Using TorchDEQ, we build a ``DEQ Zoo'' that supports six published implicit models across different domains. By developing a joint framework that incorporates the best practice across all models, we have substantially improved the performance, training stability, and efficiency of DEQs on ten datasets across all six projects in the DEQ Zoo.
TorchDEQ and DEQ Zoo are released as \href{https://github.com/locuslab/torchdeq}{open source}.
\blfootnote{Feel free to drop an email to \href{mailto:zhengyanggeng@gmail.com}{Zhengyang Geng}.}
\end{abstract}

\section{Introduction}

Deep Equilibrium Models \cite{DEQ}, or DEQs, are a recently-developed class of implicit neural network \cite{Kolter2020,El2021IDL}.  Unlike traditional feedforward models, which compute their output using a fixed-size computational graph, DEQ models define their output as a \emph{fixed point} of nonlinear systems, \ie
\begin{equation*}
\label{eq:deq}
\zstar = \f(\zstar, \x).
\end{equation*}
where $\x$ denotes the input to the network and $\zstar$ is its output. There are several notable benefits to this formulation: DEQs can be interpreted as ``infinite depth'' limits of the fixed point iteration $\z^{l+1} = \f(\z^{l}, \x)$, thus offering rich representations using relatively few parameters; they require only specifying a ``single'' layer in their architectural design; they can be trained with substantially less memory, as only the final fixed point $\zstar$ needs to be stored for backpropagation; they (often) recover path-independent solutions where the final output $\zstar$ is independent of its initialization \cite{path-independence}; and finally, they intuitively allow for a separation between the ``definition'' of a network and the ``solver'' that computes the fixed point, a separation that mirrors many settings in \eg differential equation solvers or optimizers.

Unfortunately, DEQ models in practice are often difficult to train and challenging to deploy. 
Training DEQs can often result in unstable systems, and methods for addressing these stability challenges are spread across different papers in the literature \cite{DEQ_JR,deq-flow}; similarly, backpropagation in these networks can be done in many different manners, \ie through unrolling, through implicit differentiation~\cite{DEQ}, or via inexact ``phantom'' gradients \cite{PhantomGrad}; finally, the choice of architecture and equilibrium solver must often be made anew for different applications. These challenges, we believe, have substantially limited the impact of DEQs broadly within deep learning.

To this end, we develop a modular library in this paper, dubbed TorchDEQ. TorchDEQ is a carefully designed, fully featured, and PyTorch~\cite{pythorch} based library for building and deploying DEQs. It provides decoupled and structured interfaces that allow users to customize their own general-purpose DEQs, for arbitrary tasks, through a minimal amount of code. The library supports a number of different forward solvers, backward pass methods, normalization, and regularization approaches, implementing the best practices of the entire field.

As an illustration of the library, and as a contribution in its own right, we also build a model zoo for DEQs, called the ``DEQ Zoo''. We implement six published implicit models via TorchDEQ, including DEQ Transformer~\cite{DEQ}, Multiscale Deep Equilibrium Models (MDEQ)~\cite{MDEQ}, Implicit Graph Neural Networks (IGNN)~\cite{IGNN}, Deep Equilibrium Optical Flow Estimator (DEQ-Flow)~\cite{deq-flow}, Implicit Layers for Implicit Representations (DEQ-INR)~\cite{impsq}, and Deep Equilibrium Approaches to Diffusion Models (DEQ-DDIM)~\cite{deq-ddim}.  Tellingly, when implemented using the best practices of TorchDEQ, \emph{we obtain uniformly better results, in terms of performance, stability, and efficiency, for all these models over what was reported in the original papers and the released code.}

TorchDEQ and DEQ Zoo are released alongside this paper and will be maintained in a long run. We believe this library and the insights provided by this work will become a positive step toward building a flourishing DEQ community, advancing its further studies, and empowering other machine learning areas using DEQs.

\section{TorchDEQ}

In this section, we present our TorchDEQ library and supported features in TorchDEQ. We start by briefly reviewing deep equilibrium models~\cite{DEQ}. Then we introduce the interface of TorchDEQ using sample code. We also dive into TorchDEQ to understand its computational graph design, highlighting different approaches for approximating the backward passes as well as other popular DEQ strategies.

\subsection{Intro to DEQs}

Given the input data pair $(\x, \y)$ and a loss function $\L$, DEQ is an implicit mapping from the input injection $\u(\x)$ to the fixed points $\zstar$ of a neural network $\f$. The training objective is as follows,
\begin{equation}
\begin{array}{ll}
\underset{\boldsymbol{\theta}}{\argmin} &  \L(\y, \y(\zstar)) \\
\text { s.t.} & \zstar = \f(\zstar, \u(\x))  \\
\end{array}
\end{equation}
where $\u(\x)$ is an injection function, and $\y(\zstar)$ is a decoder to produce the model prediction. In the forward pass, the ``infinite-depth'' equilibrium representation $\zstar$ can be solved by a black-box solver, \eg fixed point iteration, Anderson acceleration~\cite{anderson1965}, or Broyden's method~\cite{broyden1965class}.
Despite these ``infinite layers'',  differentiating through this fixed point system has an elegant solution. 
\begin{theorem}
By the Implicit Function Theorem (IFT)~\cite{krantz2012implicit,DEQ}, under mild conditions, the gradient of DEQ can be expressed as
\begin{align}
\label{eq:ift}
    \IFT = \underbrace{\dLdz \invJ}_{\g^\top} \frac{\partial \f(\zstar, \mathbf{x})}{\partial \theta}.
\end{align}
\end{theorem}
\vspace{-0.2cm}
This solution entails solving another ``mirror'' linear fixed point system in the backward pass to obtain the gradient $\g$. 
\begin{align}
\label{eq:ift-g}
    \g^\top = \g^\top \dfdz + \dLdz.
\end{align}
This backward equilibrium system is itself a (linear) fixed-point operation, and thus can be using similar (or even simpler) techniques as the forward pass.  Thus, we can differentiate through DEQ using $\gO(1)$ memory complexity (i.e., independent of the number of solver steps) without storing function $\f$ activations or the computational graph of the black-box solver.

\subsection{Sample Code \& Interface}

There is a commonality to these aspects listed above: in all cases, the primary attributes of deep equilibrium models are agnostic to the particular choice of function $\f$. That is to say: for different functional designs, single-variate, multi-variate, or even multi-resolution equilibrium systems, we can build a unified and modular interface for implementing DEQs.  However, implementing a DEQ is still challenging today, as all the components shown above, and further extensions, require skilled design and verification. Differences in implementation can significantly impact downstream performance, stability, and efficiency, as discussed later.

\begin{wrapfigure}{r}{0.45\textwidth}
\includegraphics[width=0.45\textwidth]{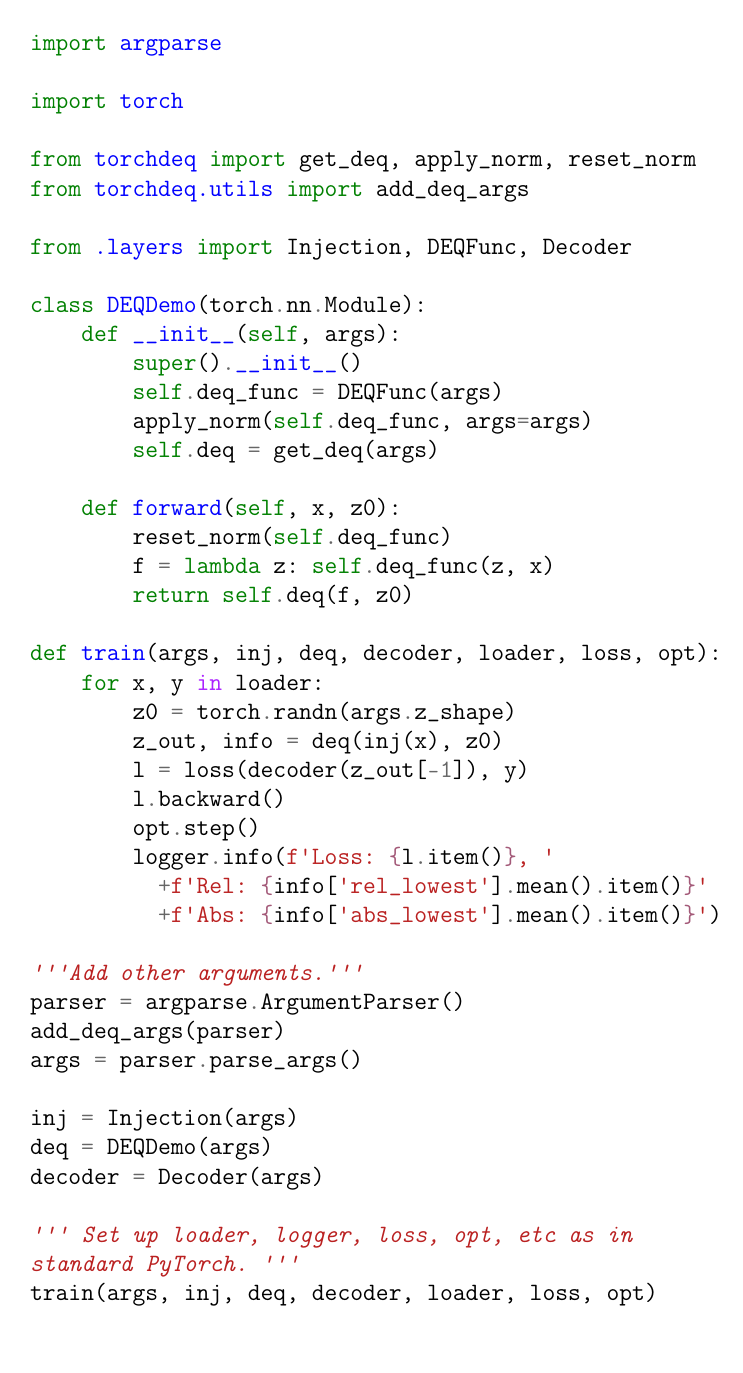}
\vspace{-0.7cm}
\caption{A mini-DEQ using TorchDEQ.}
\label{code:torchdeq}
\vspace{-1.4cm}
\end{wrapfigure}

We now describe how to build a DEQ model and its training loop using our out-of-the-box TorchDEQ library. Sample code is shown in Figure \ref{code:torchdeq}, and we describe the key functions below.

\texttt{get\_deq}: Return the DEQ solver as a Pytorch Module~\cite{pythorch}. Users need to pass a functor \texttt{f} that defines the function call to $\f$ with the input injection $\x$ and the initialization $\z^0$ for fixed point solvers. Fixed point reuse~\cite{impsq,deq-flow} can be easily done through user-chosen previous fixed points. For a multi-variate equilibrium system of different tensor shapes, like $\zstar = [\hstar, \cstar]$, one only needs to rewrite the functor with a trivial adjustment; TorchDEQ will accomplish the remaining adjustment for gradients and solvers.

\texttt{\footnotesize f = lambda h, c: self.deq\_func((h, c), x)}

\texttt{apply\_norm, reset\_norm}: Automatically apply normalizations to weight tensors in the equilibrium module $\f$. Recompute the values for each weight tensor before the next training step. See more discussions in \cref{sec:norm}.

\texttt{add\_deq\_args}: We provide a decorator for the commonly used Python argument parser. Our users can simply call \texttt{add\_deq\_args(parser)} and customize DEQs' behavior through the command line. 
This design is widely adopted by community-trusted libraries like \texttt{fairseq}~\cite{fairseq} and \texttt{timm}~\cite{timm}.

\texttt{\footnotesize python train.py ---ift ---f\_solver anderson ---b\_solver broyden}

For example, the above command launches the training using implicit differentiation as the backward, Anderson Acceleration as the forward solver, and Broyden's method as the backward solver. More command options will be introduced along with the DEQ features in TorchDEQ.

TorchDEQ's compact and modular interfaces enable users to focus on how to abstract, formulate and define their demands as an equilibrium model $\f$ and devise its interaction with other explicit layers like injection and decoder.
The modular design of TorchDEQ creates an ``abstraction'' for DEQs and reduces the cost of learning, implementing, and tuning DEQs to a minimum.
In the following sections, we introduce the features of TorchDEQ and their control command. 

\vspace{-0.2cm}
\subsection{Backward Pass}
\label{sec:backward}

TorchDEQ internally creates computational graphs for solvers and gradients. Users will receive a group of tensors registered with gradients. 
Users work on the outputs of the implicit model just as they would explicit layers and tensors.  However, when computing gradients, we transparently compute the backward pass using specialized methods.  We support two types of backward passes, namely implicit differentiation (IFT)~\cite{DEQ} and phantom gradients (PG)~\cite{PhantomGrad}. In practice, we find that both and their combination suffice to provide empirically appealing results within a reasonable time frame.

\tbf{Implicit Differentiation (IFT)}. Implicit differentiation is the standard approach to differentiate through fixed points. As we discussed in \cref{eq:ift-g}, implicit gradients can be solved from another linear fixed point system in the backward pass. Users can declare IFT through \texttt{---ift}, set a backward solver using \texttt{---b\_solver broyden}, and set up solver configurations like maximum solver steps \texttt{---b\_max\_iter 30} and stopping criteria \texttt{---b\_tol 0.001}. 

\tbf{Phantom Grad (PG)}. Phantom Gradient~\cite{PhantomGrad} is a structured approximation of IFT that keeps the descent direction,
\begin{equation}
\label{eq:pg}
    \IFT \approx \PhantomGrad = \dLdz \A.
\end{equation}
where $\left\langle \PhantomGrad, \IFT \right\rangle > 0$ preserves a valid gradient update, and $\A$ is an approximate Jacobian defined below.

Phantom gradients can be applied to computational graphs of various solvers, which is similar to IFT for differentiating fixed points.
The previous view considers IFT as an exact gradient and PG as an inexact gradient. However, because of the numerical errors in solving the forward and IFT, we do not further distinguish from exact or inexact gradients and call them backward passes together instead.

An instantiation of PG used in different implicit models is to unroll the equilibrium module $\f$ over the solved (approximate) fixed points $\z^p$ with a damping factor $\tau$,
\begin{equation}
\label{eq:upg-forward}
  \z^{p+1} = \tau \f(\z^p) + (1 - \tau) \z^p,
\end{equation}
which defines the following $\A$ matrix,
\begin{equation}
\label{eq:upg}
  \A = \tau \sum_{k=p}^{K-1}\prod_{s=k+1}^{K-1} \left( \left. \tau \frac{\partial \f}{\partial \z} \right|_{\z^{s}} + \left( 1 - \tau \right) \I \right) \left. \frac{\partial \f}{\partial \theta} \right|_{\z^{k}}.
\end{equation}
Users can call PG, for example, by \texttt{---grad 5 ---tau 0.6} combined with Broyden's method as the forward solver.

We do not define separate support for backpropagation through time (BPTT)~\cite{werbos1990backpropagation} and its truncated version  because they can be expressed as special cases of PG given $\tau=1.0$ and removing the forward solver, \ie the solver and gradient are solely defined by an unrolled process of $\f$. Using a command of \texttt{---f\_max\_iter 0 ---grad 12 ---tau 1.0} defines a computational graph of BPTT-12.

We also provide an interface \texttt{mem\_gc} to reduce the memory complexity of any unrolled computational graph to $\mathcal{O}(1)$ \wrt the equilibrium function $\f$ activations via gradient checkpointing~\cite{gradckpt}. As an alternative to IFT and PG, users can trade training time by 1.5$\times$ to obtain a much lower memory overhead through this interface.

\subsection{Solvers}
\label{sec:solver}

Previous DEQ projects usually wrote their own fixed point solvers. It is unsurprising that these task-dependent solvers can be inefficient or even sometimes unreliable when applied to different domains. 
To tackle this problem, we implement, verify, and polish our solver implementations for TorchDEQ.
We especially optimize the batching and memory access for multi-variate systems.
Our solvers are reliable in various tasks, robust across different settings, and agnostic to the scale of fixed point equations and their tensor shapes.
These efforts lead to impressive efficiency improvements over multi-variate and multi-scale equilibrium systems.

In TorchDEQ, we support the following solvers. To call these solvers, users can type \texttt{---f\_solver} or \texttt{---b\_solver} with solver names in the command line and their maximum iterations \texttt{---f\_max\_iter 20} and stopping criteria \texttt{---f\_tol 1e-2}. In addition, keyword arguments can be passed to the DEQ class to tune a solver. An example of customizing Anderson Acceleration~\cite{anderson1965}, for instance, could be accomplished as

\texttt{\footnotesize z\_out, info = self.deq\_solver(f, z, solver\_kwargs=\{'tau': 0.8, 'm':6\})}. 

\tbf{Fixed Point Iteration}. Fixed point iteration is the classic solver for solving fixed points $\zstar$, described by the following numerical scheme,
\begin{equation}
\label{eq:naive}
\z^{k+1} = \f(\z^k).
\end{equation}
Its convergence can be guaranteed by a bounded Jacobian spectral radius of $\f$. Users can type \texttt{---f\_solver fixed\_point\_iter} to call fixed point iterations.

\tbf{Anderson Solver}. Anderson Acceleration, or Anderson mixing~\cite{anderson1965}\footnote{This is also called Type-I Anderson Acceleration.}, is an acceleration technique for fixed point iterations using the linear combination of past $m+1$ fix points estimations. Its update employs this numerical scheme,
\begin{equation}
\label{eq:anderson}
\z^{k+1} = \tau \sum_{i}^m \alpha^k_i \f(\z^{k-m+i}) + (1 - \tau) \alpha^k_i \z^{k-m+i}, \\
\end{equation}
where $\tau$ is a damping factor with a default value $1.0$. Given  $\gf(\z) = \f(\z) - \z$, $\rmG^k = [\gf(\z^{k-m}), \cdots, \gf(\z^k)]$, $\boldsymbol{\alpha}^k = [\alpha^k_0, \cdots, \alpha^k_m]$ is solved from
\begin{equation}
\begin{array}{ll}
\underset{\boldsymbol{\alpha}}{\argmin} &  \| \rmG^k \boldsymbol{\alpha} \|_2 \\
\text { s.t.} & \mathbf{1}^\top \boldsymbol{\alpha} = 1 \\
\end{array}
\end{equation}
Users can call Anderson Acceleration via \texttt{---f\_solver anderson} and tune it referring to the sample above.

\tbf{Broyden Solver}. Broyden's method~\cite{broyden1965class} is a quasi-Newton solver for fixed point equations. By maintaining a buffer, Broyden's method approximates the Jacobian inverse in Newton's method through low-rank updates,
\begin{equation}
\label{eq:broyden}
\z^{k+1} = \z^{k} - \alpha \cdot \B^{k}\gf(\z^k),
\end{equation}
where $\B^{k}$ is the approximation of Jacobian inverse $\rmJ_{\gf}$ using $\Delta \z^k = \z^{k} - \z^{k-1}$ and $\Delta \g^k = \gf(\z^k) - \gf(\z^{k-1})$,
\begin{equation}
\label{eq:invj}
\B^{k} = \B^{k-1} +\frac{\Delta \z^k - \B^{k-1} \Delta \g^k}{\Delta {\z^k}^\top \B^{k-1} \Delta \g^k} \Delta {\z^k}^\top \B^{k-1}.
\end{equation}

\cref{eq:broyden} can be written into a matrix-vector product that further avoids storing $\B^k$ in memory,
\begin{equation}
\label{eq:broyden-mvp}
\z^{k+1} = \z^{k} - \alpha \cdot (\B^0 + \rmU^k{\rmV^k}^\top)\gf(\z^k),
\end{equation}
where $\rmU^k$ and $\rmV^k$ represent $m$ past estimations for the low-rank approximation via the \textit{Sherman-Morrison formula}~\cite{sherman1950adjustment}. Users can call the Broyden's method through \texttt{---f\_solver broyden} or its limited-memory version, for example, by setting \texttt{solver\_kwargs=\{'l\_thres':$m$\}}.

\subsection{Normalization}
\label{sec:norm}
Normalization techniques are vital to modern deep equilibrium models. Unlike popular normalization methods~\cite{BN,LN} applied to \textit{representations}, DEQs additionally rely on normalization for \textit{weight} tensors, \eg Weight Normalization~\cite{WN}, 
Spectral Normalization~\cite{SN}, which we support their DEQ versions in TorchDEQ.

Normalization significantly smooths the fixed-point landscape of $\z$ given the input data $\x$, and makes the fixed points easier to solve in practice. We find this effect is usually underestimated in prior literature.

For a weight matrix $\W \in \Rdim{m \times n}$, Weight Normalization (WN) parameterizes the weight into
\begin{equation}
\label{eq:wn}
\W_{i:} = \W_{i:} \frac{\rvg_i}{ \| \W_{i:} \|},
\end{equation}
where $\| \cdot \|$ stands for vector $L_2$ norm, $\g$ is a learnable scaling factor, while Spectral Normalization (SN) states
\begin{equation}
\label{eq:sn}
\W  = \frac{ \W }{ \| \W \|_2} = \W \frac{1}{ \| \W \|_2}  
\end{equation}
where $\| \cdot \|_2$ is the spectral norm, usually computed by power iterations~\cite{SN}.

In TorchDEQ, we support both normalization methods via the formalism,
\begin{equation}
\label{eq:norm}
\W  = \W  \circ \rvf = \W \circ \frac{\rvg}{ \gN(\W) }  
\end{equation}
where $\circ$ is the row-wise multiplication, and $\gN$ stands for computing the relevant ``norm'' for the weight matrix. Following WN, we add a learnable scaling $\g$ to DEQ SN, as we find that it enables SN to match WN's generalization performance on DEQ-Flow~\cite{deq-flow}.

Inspired by gradient clipping~\cite{gradclip}, we also introduce an operation that significantly stabilizes the training of implicit graph neural networks~\cite{IGNN} on node classification, \ie by clipping the rescaling factor $\rvf$ to a threshold $t$,
\begin{equation}
\label{eq:deq-norm}
\W  = \W  \circ \min(t, \rvf) = \W \circ \min(t, \frac{\rvg}{ \gN(\W) }).
\end{equation}
In TorchDEQ, this can be enabled by \texttt{---norm\_clip} with \texttt{---norm\_clip\_value $t$}.

Classic implementations for WN and SN reset the weight in every \texttt{forward} call. However, this is a visible waste for DEQ because the equilibrium module $\f$ will be called many times until convergence. Then the same weight parameterizations are applied by the number of $\f$ function calls. 
Plus, decorating normalization has to be manually coded for each module using the vanilla PyTorch implementation.

Instead, in TorchDEQ, We provide unified interfaces for automatically decorating the entire equilibrium module $\f$ through \texttt{apply\_norm} (with the keyword argument \texttt{filter\_out} to skip some modules) and \texttt{reset\_norm} for resetting without wasting compute. 
After training, the normalization decorations can be removed by \texttt{remove\_norm}, as they do not change the model but ease its training.

Users can specify \texttt{---norm\_type weight\_norm} for WN, \texttt{---norm\_type spectral\_norm} for SN, and additionally \texttt{---norm\_no\_scale} for removing the learnable scaling $\g$.

\subsection{Regularization} 
There is an ongoing conceptual change in the modern interpretations of DEQ learning.
Instead of considering DEQ models just as learning an implicit fixed point mapping $\x \rightarrow \zstar$, they are thought of as learning the equilibrium \emph{landscape} $\x \rightarrow \gB(\z)$~\cite{DEQ_JR,deq-flow,swami2021joint} that contains a unique and performant fixed point $\zstar$~\cite{path-independence}. 

The regularity~\cite{DEQ_JR} of the equilibrium module $\f$ guarantees a fast convergence to fixed points $\zstar$ despite using a simple solver. The correspondence between the equilibrium landscape $\gB(\z)$ and loss landscape~\cite{deq-flow} indicates a strong correlation between fixed point errors $\| \f(\z) - \z \|$ and the losses $\L(\y(\z))$. The path independence~\cite{path-independence}, \ie converging to the steady state regardless of initialization, allows us to exploit test time computation better. Altogether, these works consider DEQs as a dynamic implicit neural network that can obtain strong results in the early equilibrium-solving process and gradually improve its prediction as approaching fixed points $\zstar$. 
In TorchDEQ, we support techniques promoting these DEQ properties and the regularity of equilibrium landscapes.

\tbf{Jacobian Regularization (JR)}. Jacobian Regularization~\cite{DEQ_JR} penalizes the upper bound of Jacobian spectral radius $\rho(\Jf)$. 
\begin{equation}
\label{eq:jr}
    \rho(J_{\f}) \leq \| J_{\f} \|_F = \sqrt{\text{tr}(J_{\f}^\top J_{\f})} 
\end{equation}
Computationally, this is accomplished by adding a loss term using the stochastic Hutchinson trace estimator~\cite{Hutchinson}, \eg sampling $\epsilon$ from a standard Gaussian,
\begin{equation}
    \text{tr}(J_{\f}^\top J_{\f})  \approx \sum_{\epsilon\sim p(\epsilon)} \| J_{\f} \epsilon \|_2^2
\end{equation}
In TorchDEQ, we offer an interface \texttt{jac\_reg} that takes $\f(\z)$ and $\z$ to compute the JR loss.

\tbf{Fixed Point Correction (FC)}.
\label{sec:fc}
Fixed point correction~\cite{deq-flow} helps learn a smooth equilibrium landscape by regularizing intermediate states from the fixed points solving process. Considering a sequence $\Tilde{\Z} = [\z^{k_1}, \cdots, \z^{k_n}]$ that converges to $\zstar$, correction can either decode the states and supervise the predictions,
\begin{equation}
     \min \sum_i^n \gamma^{n-i} \L(\y, \y(\z^{k_i})), \gamma \leq 1,
\end{equation}
or apply Jacobian regularization to this sequence~\cite{swami2021joint},
\begin{equation}
     \min \L(\y, \y(\zstar)) + \gamma \sum_i^n \sum_{\epsilon\sim p(\epsilon)} \| J_{\f}(\z^{k_i}) \epsilon \|_2^2.
\end{equation}
Implementation-wise, we need to create this sequence $\Tilde{\Z}$ via TorchDEQ. We support two types of commands; \texttt{---indexing 20 30} can sample, for example, states $k_i=20,30$, while \texttt{---n\_states $n$} uniformly sample $n$ states from a solver. The resulting \texttt{z\_out} from the DEQ class is thus a \texttt{list} object that contains this $\Tilde{\Z}$ sequence.
TorchDEQ can naturally and internally handle backward passes for this sequence, differentiating the best fixed point estimation $\z^{k_n}$ via IFT or PG and all the other states through PG.

\tbf{Random Iterations}. A randomized maximum number of function calls can act as a positive intervention for path independence~\cite{path-independence}. In TorchDEQ, users can pass a random \texttt{f\_max\_iter} to the DEQ class at each training iteration to achieve this regularization.

\tbf{Mixed Initialization}. Mixed initialization~\cite{path-independence} adopts half zeros and half standard Gaussian to reduce the dependence on initialization. In TorchDEQ, users can call \texttt{mixed\_init} with a target shape to initialize the fixed point solvers.

\section{DEQ-Zoo}
\label{sec:zoo}

Empowered by TorchDEQ, we establish DEQ Zoo to systematically revisit, support, and make these DEQ ideas broadly available to the research community. The reliability of TorchDEQ is verified on six published DEQ projects, showing improved performance, stability, and efficiency. We present DEQ~\cite{DEQ}, MDEQ~\cite{MDEQ}, IGNN~\cite{IGNN}, DEQ-Flow~\cite{deq-flow}, DEQ-INR~\cite{impsq}, and DEQ-DDIM~\cite{deq-ddim} in the main paper, while the license information and additional efficiency analysis are left to the Appendix.

\begin{table}[t]
\footnotesize
\begin{minipage}{0.49\textwidth}
    \centering
    \caption{\tbf{Performance, Speed, Memory of DEQ Transformer.} We report the Perplexity (PPL) (lower is better) on WikiText-103. Rel indicates relative fixed point errors. Time is measured by the training time used. $^\dagger$ corresponds to the results from the original published paper. TorchDEQ improves DEQ transformers to much better efficiency, performance, and stability.}
    \resizebox{\textwidth}{!}{
        \begin{tabular}{@{}lcccc@{}}
        \toprule
        Method               &  PPL  & Rel & Time  & Memory \\
        \midrule    
        Original DEQ$^\dagger$ &  24.0    &  0.10      & 1.00$\times$      & 30.5GB \\
        DEQ~(Ours, IFT)        &  23.8    &  0.10      & 0.98$\times$      & 29.5GB\\
        DEQ~(Ours, IFT+JR)     &  23.7    &  0.001     & 1.02$\times$      & 32.8GB \\
        \midrule  
        DEQ~(Ours, Final)      &  \tbf{22.4}    &  0.10      & \tbf{0.36$\times$}  & \tbf{27.8GB}\\
        \bottomrule
        \end{tabular}
    }
    \label{Tab:deq}
\end{minipage}
\hspace{0.02in}
\begin{minipage}{0.49\textwidth}
    \centering
    \caption{\tbf{Stability of IGNN.} We report the Macro-F1 (\%) (higher is better) on the PPI dataset. $^\dagger$ corresponds to the results in the original published paper. $^\ddagger$ stands for the results in \cite{PhantomGrad} using phantom gradients. We test our implementation using IFT and PG, denoted as (Ours). TorchDEQ significantly helps stabilize the training dynamics of IGNN.} 
    \vspace{0.025in}
    \resizebox{0.535\textwidth}{!}{
        \setlength{\tabcolsep}{1.75mm}
        \begin{tabular}{@{}lcc@{}}
        \toprule
        Method                   & Median & Best               \\
        \midrule
        IGNN$^\dagger$~(IFT)     &  72.7  & 97.9               \\
        IGNN~(Ours)              &  97.8  & 98.2               \\
        \midrule
        IGNN$^\ddagger$~(PG)     &  93.9  & 98.2               \\
        IGNN~(Ours)              &  \tbf{98.0}  & \tbf{98.6}   \\
        \bottomrule
        \end{tabular}
    }\hspace{-0.02in}
    \resizebox{0.46\textwidth}{!}{
        \setlength{\tabcolsep}{1.3mm}
        \begin{tabular}{@{}lc@{}}
        \toprule
        Method      &  Macro-F1     \\
        \midrule
        GCN         & 59.2          \\
        GraphSAGE   & 78.6          \\
        GAT         & 97.3          \\
        \midrule
        IGNN~(Ours) & \tbf{98.0}   \\
        \bottomrule
        \end{tabular}
    }
    \label{Tab:ignn}
\end{minipage}
\vspace{-0.5cm}
\end{table}
\subsection{DEQ}
The first Deep Equilibrium Model~\cite{DEQ} is a sequence model that takes advantage of transformers~\cite{transformer,trans-xl} in its model design. Given the injection $U(\x_{0:T})$ from the input sequence and the past context $\zstar_{0:t}$, DEQ transformer predicts the next tokens via the fixed points $\zstar_{t:T}$ of a transformer block,
\begin{equation}
\label{eq:deq-seq}
\begin{array}{llll}
& \rvq, \rvk, \rvv & = & \W \zstar_{0:T} + U(\x_{0:T}) \\
& \Tilde{\z}       & = & \zstar_{t:T} + \text{Attention}\left(\rvq, \rvk, \rvv\right)   \\ 
& \zstar_{t:T}     & = & \Tilde{\z}   + \text{FFN}\left(\Tilde{\z} \right)  \\
\end{array} 
\end{equation}
where Attention is MultiHead Decoder Attention~\cite{transformer}, FFN is a 2-layer feed-forward network.

In DEQ Zoo, we implement the DEQ transformer and benchmark it through the word-level language modeling on WikiText-103~\cite{wiki}. The model details and training protocols are redesigned based on TorchDEQ. These efforts contribute to substantially better long-term dependency modeling, training time, and even memory usage, as shown in \cref{Tab:deq}.
\begin{table}[ht]
\centering
\caption{\tbf{Training Speed of MDEQ.} We report the Top-1 accuracy (\%) (lower is better) on different datasets. $^\dagger$ corresponds to the results in the original published paper. $^\ddagger$ represents results in phantom gradients~\cite{PhantomGrad}.  TorchDEQ accelerates the training of multi-variate equilibrium systems.}
\vspace{0.1in}
\begin{tabular}{@{}lcccc@{}}
\toprule
Method &  Dataset & Params &  Median & Time \\
\midrule    
MDEQ-Tiny$^\dagger$   & CIFAR10  & 159K &  85.1 & 1.00$\times$ \\
MDEQ-Tiny$^\ddagger$  & CIFAR10  & 159K &  85.7 & 0.43$\times$ \\
MDEQ-Tiny~(Ours)      & CIFAR10  & 159K &  85.7 & 0.26$\times$ \\
\midrule 
MDEQ-Large$^\dagger$  & CIFAR10  & 10M &  93.8  & 1.00$\times$ \\
MDEQ-Large$^\ddagger$ & CIFAR10  & 10M &  95.0  & 0.63$\times$ \\
MDEQ-Large~(Ours)     & CIFAR10  & 10M &  94.8  & 0.37$\times$  \\
\midrule 
MDEQ$^\ddagger$       & ImageNet & 18M &  75.7  & 1.00$\times$ \\
MDEQ~(Ours)           & ImageNet & 18M &  75.7  & 0.60$\times$ \\
\bottomrule
\end{tabular}
\vspace{-0.2cm}
\label{Tab:mdeq}
\end{table}
\subsection{MDEQ}
When DEQ was proposed, doubt was cast toward DEQ, \ie whether this idea could scale up to high-resolution computer vision tasks. Multiscale Deep Equilibrium Models (MDEQ)~\cite{MDEQ} first demonstrated this possibility.
MDEQ solves a joint equilibrium of multi-resolution features, $\zstar = [\zstar_1, \cdots, \zstar_n]$. For each feature scale, MDEQ follows this update scheme,
\begin{equation}
\label{eq:mdeq}
\begin{array}{llll}
& \breve{\z}_i  & = & \text{GN}(\text{Conv}(\zstar_i))   \\
& \Tilde{\z}_i  & = & \text{GN}(\text{Conv}(\text{ReLU}(\breve{\z}_i)) + \textbf{1}_{i=1} \cdot U(\x))   \\
& \bar{\z}_i    & = & \text{GN}(\text{ReLU}(\Tilde{\z}_i + \zstar_i))        \\
& \zstar_i      & = & \text{GN}(\text{Conv}(\bar{\z}_i + \sum_{j \neq i} \text{Resize}(\bar{\z}_j)))        \\
\end{array} 
\end{equation}
where GN stands for Group Normalization~\cite{GN}, $\textbf{1}_{i=1}$ is an indicator that injects input data into the largest resolution $i=1$, and Resize means downsampling for $j<i$ and upsampling for $j>i$.

TorchDEQ can accelerate multi-variate equilibrium systems and demonstrate over $30\%$ time saving compared to an accelerated baseline~\cite{PhantomGrad}, using the same configuration for Broyden solver. Benefiting from the smooth equilibrium landscape, we can train and solve MDEQ using fixed point iteration on ImageNet~\cite{ILSVRC15}, which leads to a $40\%$ training time reduction. (See \cref{Tab:mdeq})

\subsection{IGNN}
Implicit Graph Neural Network (IGNN)~\cite{IGNN} is the first implicit model in the graph domain. It solves the following equilibrium graph features $\Zstar \in \Rdim{n \times d}$,
\begin{equation}
\label{eq:ignn}
\Zstar  =  \phi\left(\A \Zstar \W + U(\X) \right)  
\end{equation}
where $\A \in \Rdim{n \times n}$ is the adjacency matrix of input graph. IGNN presents a tighter theoretical analysis of the well-posedness of implicit models~\cite{IGNN,El2021IDL}. Its well-posedness is guaranteed by the \textit{Perron-Frobenius (PF) eigenvalue} $\lambda_{pf}(| \A \otimes \W |) \le 1$, where $\otimes$ stands for the Kronecker product. In practice, this condition is achieved by a projected gradient descent over $\W$.

Through TorchDEQ, our IGNN becomes considerably more stable using the clipped DEQ SN (with an absolute Macro-F1 gain of over $4\%$, see \cref{Tab:ignn}), outperforms explicit graph models~\cite{kipf2017gcn,ppi,veličković2018graph}, and matches the performance of recent implicit graph networks~\cite{liu2021eignn,chen2022gind} on multi-label node classification.
\begin{table}[ht]
\vspace{-0.2in}
\centering
\caption{\tbf{Performance of DEQ-Flow.} We report the Average End Point Error~(AEPE), and F1-all (\%) (lower is better) on Sintel and KITTI 2015 datasets. $^\dagger$ corresponds to the results in the original published paper. $^\circ$ indicates results by fixed point correction. The bold font stands for the best result, and the underlined results rank 2nd. TorchDEQ significantly improves the performance of DEQ-Flow.}
\vspace{0.1in}
\footnotesize
\begin{tabular}{@{}lcccc@{}}
\toprule
\multirow{2}{*}{Method} & \multicolumn{2}{c}{Sintel (train)} & \multicolumn{2}{c}{KITTI-15 (train)} \\
\cmidrule(lr){2-3}
\cmidrule(lr){4-5}
& Clean & Final & AEPE & F1-all \\
\midrule    
RAFT~\cite{RAFT}                   & 1.43        & \tbf{2.71}  & 5.04       & 17.4       \\
DEQ-Flow-B$^\dagger$               & 1.48        & 2.81        & 5.01       & 16.3       \\
DEQ-Flow-B~(Ours)                  & \tbf{1.42}  & 2.75        & \tbf{4.60} & \tbf{15.0} \\
\cmidrule[\lightrulewidth](r{0.3em}){1-5}
RAFT-H                             & 1.36	& 2.59	&  4.47	 & 
16.2  \\
DEQ-Flow-H$^\dagger$               & 1.41       & 2.75   
    &  4.38      & 14.9      \\
DEQ-Flow-H$^{\dagger \circ}$       & 1.34       & 2.60        & 3.99  & 13.5           \\
DEQ-Flow-H~(Ours)                  & \tbf{1.28} & \tbf{2.58}  & \tbf{3.77} & \tbf{13.0}  \\
\cmidrule[\lightrulewidth](r{0.3em}){1-5}
RAFT~\cite{RAFT}                    & 1.43  & 2.71        & 5.04            & 17.4       \\
GMA~\cite{GMA}                      & 1.30  & 2.74        & 4.69            & 17.1       \\
SeperableFlow~\cite{SepFlow}        & 1.30  & 2.59        & 4.60       & 15.9            \\
CRAFT~\cite{craft}                  & 1.27  & 2.79        & 4.88       & 17.5            \\
KPA-Flow~\cite{kpa-flow}            & 1.28  & 2.68        & 4.46       & 15.9            \\
GMFlow~\cite{gmflow}                & \ul{1.09}  & \ul{2.48}   & 7.77       & 23.4       \\
FlowFormer~\cite{flowformer}        & \tbf{1.01}  & \tbf{2.40} & 4.09       & 14.7      \\
DEQ-Flow-H$^{\dagger \circ}$        & 1.34  & 2.60             & \ul{3.99}  & \ul{13.5}  \\
DEQ-Flow-H~(Ours, $3\times$Iters)   & 1.27  & \ul{2.48}        & \tbf{3.78} & \tbf{13.4} \\
\bottomrule
\end{tabular}
\vspace{-0.2cm}
\label{Tab:deq-flow}
\end{table}
\subsection{DEQ-Flow}
Deep Equilibrium Optical Flow Estimator (DEQ-Flow)~\cite{deq-flow} is an industry-grade application of DEQ on optical flow estimation, where DEQ-Flow solves the hidden equilibrium $\hstar \in \Rdim{C \times H\times W }$ and the per-pixel correspondence $\fstar \in \Rdim{2 \times H \times W}$, \ie optical flow, between two consecutive frames,
\begin{equation}
\label{eq:deq-flow}
\begin{array}{llll}
& \x          & = & \text{Conv}\left([\q,\, \fstar, \gC(\fstar+\c^0)]\right) \\[0.3mm]
& \hstar      & = & \text{ConvGRU}\left(\hstar, [\x,\, \q] \right)   \\[0.15mm]
& \fstar      & = & \fstar + \text{Conv}\left(\hstar \right)        \\
\end{array} 
\end{equation}
where $\q$ is the query embedding of the first frame, $\gC(\fstar+\c^0)$ stands for the ``correlation lookup''~\cite{RAFT} using the flow estimation.
Notably, optical flow data are hard to collect and label. State-of-the-art flow estimators have trained on synthetic datasets~\cite{chairs,things} and tested on animated datasets and real-world data with large motions, imposing huge challenges to their out-of-distribution (OOD) generalization.

Amazingly, DEQ-Flow presents ultra-strong OOD generalization abilities in large-scale real-world challenges. Our DEQ-Flow ranks $1_{st}$ on the real-world KITTI dataset~\cite{kitti} and co-ranks $2_{nd}$ on the Sintel Final split~\cite{sintel} (w/ blurs and motion effects) under this training pipeline (See \cref{Tab:deq-flow}), despite only using half the training iterations compared to the latest transformer-based flow models~\cite{gmflow,gmflow+}.

\subsection{$\text{Implicit}^2$ (DEQ-INR)}

Implicit Neural Representations~\cite{park2019deepsdf,siren,tancik2020fourier,mildenhall2021nerf} learns low-dimensional mappings from input grids to attributes like colors, \ie $\x = (x, y) \in \Rdim{2} \rightarrow \y = (R, G, B) \in \Rdim{3}$. 
It allows for storing and compressing continuous representations for images or 3D scenes in a neural net.

\citet{impsq}~(DEQ-INR) exploit the parameter efficiency of DEQ to build equilibrium implicit neural representations using SIREN~\cite{siren} or multiplicative filter networks (MFN)~\cite{mfn},
\begin{align}
\zstar      & = \text{Sin}\left(\W\zstar + U(\x) + \rvb \right)   \label{eq:deq-siren} \\
\zstar      & = (\W \zstar + \rvb) \circ \text{Filter}\left(U(\x) \right) \label{eq:deq-mfn}
\end{align}
where $\text{Sin}$ is the sinusoidal activation function, and Filter represents the Fourier filter or Gabor filter.

\begin{table}[h]
\centering
\caption{\tbf{Evaluation of DEQ-INR on image generalization, video representation, and audio representation.} We report the Peak signal-to-noise ratio~(PSNR) (higher is better). $^\dagger$ corresponds to the results in the original published paper.}
\vspace{0.1in}
\begin{tabular}{@{}lcccc@{}}
\toprule
Method &  Nature & Text & Video & Audio\\
\midrule    
SIREN~\cite{siren}                & 25.17        & 27.03       & 26.52        & 48.44 \\
DEQ-SIREN$^\dagger$~\cite{impsq}  & 25.60        & 27.78       & 26.61        & 48.24 \\
DEQ-SIREN (Ours)                  & \tbf{25.70}  & \tbf{28.58} & \tbf{30.37}  & \tbf{51.35}  \\
\bottomrule
\end{tabular}
\label{Tab:deq-inr}
\end{table}

In \cref{Tab:deq-inr}, we show that TorchDEQ brings remarkable improvements over the image, video, and audio domains in both reconstruction (Video, Audio) and generalization (Nature, Text) performance.

\subsection{DEQ-Diffusion Solver}

\setlength\tabcolsep{4pt}
\begin{table}[t]
\centering
\caption{\tbf{Model inversion using DEQ-DDIM.} We report the mean squared errors~(MSE) and time used for image inversion (lower is better). $^\dagger$ corresponds to the results in the original published paper.}
\vspace{0.1cm}
\begin{tabular}{@{}lcc@{}}
\toprule
Method                               & MSE      & Time                  \\
\midrule    
DDIM~\cite{ddim}                     & 15.74$\pm$8.7    & 1.00$\times$  \\
DEQ-DDIM$^\dagger$~\cite{deq-ddim}   & 0.35$\pm$0.10   & 0.27$\times$   \\
DEQ-DDIM~(Ours)                      & \tbf{0.17$\pm$0.09}  & \tbf{0.12$\times$}  \\
\bottomrule
\end{tabular}
\vspace{-0.2cm}
\label{Tab:deq-ddim}
\end{table}

Diffusion models~\cite{diffusion,ddpm,ddim} and their continuous counterparts through Ordinary Differentiation Equations (ODE)~\cite{Karras2022edm,liu2022flow} or Stochastic Differential Equations (SDE)~\cite{scoresde}, leverage a reverse diffusion process from time step $T$ to $0$ to generate a data sample. Their numerical solvers/samplers adopt the following forms,
\begin{equation}
    \z_{t-1} = a_t \z_t + b_t \unet \left(\z_t, t\right) + c_t \x_t,
    \label{eq:ddim-backward}
\end{equation}

where $\z_t$ is the data sample at time step $t$; $\unet$ is the denoising network conditioning on the time step $t$; $\x_t\sim \gN(\tbf{0}, \I)$ is the trajectory noise along the sampling process; $a_t$, $b_t$, $c_t$ are time-dependent constants.
At time step $T$, $\z_T$ is pure noise from a prior distribution, while $\z_0$ at time step $0$ is the generated clean data sample. Diverse solvers~\cite{ddim,Karras2022edm,lu2022dpm,zhang2022fast} employ different constants for ODE/SDE discretization or incorporate additional correction steps. However, they uniformly retain the element of sequential dependency.

As previously mentioned, diffusion models employ a sequential, stochastic, and time-dependent sampling process, which stands in stark contrast to the parallel, deterministic, and time-independent nature of DEQ models. For a long time, it has been uncertain how to reconcile these two distinct computational paradigms.

Recent work~\cite{deq-ddim} introduces a parallel scheme via the equilibrium of the \textit{sampling chain} to bridge the gap between two model families.
This fixed point system can be derived by first expanding \cref{eq:ddim-backward},
\begin{align*}
    \z_{t-1} 
    &= a_t a_{t+1} \z_{t+1} + a_t \left(b_{t+1} \unet \left(\z_{t+1}, t+1\right) + c_{t+1} \x_{t+1}\right) + \left(b_t \unet \left(\z_{t}, t\right) + c_t \x_t\right) \\
    &= \hdots \\
    &= \prod_{i \geq t}^T a_i \z_T + \sum_{i \geq t}^T r_i \left(b_i \unet \left(\z_i, i\right) + c_i \x_i \right),
\end{align*}
where $r_i = \prod_{i > j \geq t} a_j$ for $i > t$, and $r_t = 1$. 

Then, concatenate the trajectory of the sampling chain into a matrix form,
\begin{align}
\label{eq:diffusion-parallel}
    \begin{bmatrix}
    \z_{0}   \\
    \z_{1}   \\
    \vdots   \\ 
    \z_{T-1}
    \end{bmatrix}  &= 
    \A
    \begin{bmatrix}
    \z_{T}   \\
    \z_{T}   \\
    \vdots   \\
    \z_{T}   \\
    \end{bmatrix}
    +
    \rmR (\rmB 
        \begin{bmatrix}
        \unet(\z_1, 1)    \\
        \unet(\z_2, 2)    \\
        \vdots            \\
        \unet(\z_T, T)    \\
        \end{bmatrix} 
    + \rmC 
        \begin{bmatrix}
        \x_1             \\
        \x_2             \\
        \vdots           \\
        \x_T             \\
        \end{bmatrix} 
    )
\end{align}

where we have 
$\A_{i,i} = \prod_{j \geq i}^T a_j$, 
$\rmB_{i,i} = b_i$, 
$\rmC_{i,i} = c_i$ for $i=1,\cdots,T$, otherwise $0$; 
and $\rmR_{i,k} = \prod_{k > j \geq i} a_j$ for $k > i$, $\rmR_{i,i} = 1$, otherwise $0$.

\cref{eq:diffusion-parallel} can be solved in a ``zig-zag'' pattern, \ie from $\z_T$ (right) to $\z_{T-1}$ (left), from $\z_{T-1:T}$ (right) to $\z_{T-2}$ (left), until from $\z_{1:T}$ (right) to $\z_0$ (left), corresponding to a sequential solver. 

In addition, as $\Z$ appears on both sides of the equation, it is also a fixed point system.
Denote sampling trajectories $\Z = [\z_0, \z_1, \cdots, \z_{T-1}]$; 
initial noises $\Z_T= [\z_T, \z_T, \cdots, \z_T]$; noises $\X= [\x_1, \x_2, \cdots, \x_T]$; 
denoiser outputs $\unet (\Z, \rvt) = \left[\unet(\z_1, 1), \unet(\z_2, 2), \cdots, \unet(\z_T, T) \right]$. \cref{eq:diffusion-parallel} translates to

\vspace{-0.2cm}
\begin{equation}
\label{eq:deq-ddim}
    \Zstar = \A \Z_T + \rmR (\rmB \unet (\Zstar, \rvt) + \rmC \X),
\end{equation}

where initial state $\z_T$ and noises $\X$ are input injection; the sampling trajectory $\Zstar$ is the equilibrium of this function. It is also worth noting that any chunk $\z_{t:t+p}$ of this equation is a (sub) fixed point system. This parallel scheme produces exactly the same result as the sequential sampler once it reaches the equilibrium.

This unification between DEQ and diffusion models enables us to perform the sampling \textit{in parallel} through advanced fixed point solvers, which offers a better convergence rate, and allows for efficient differentiation through sampling, \eg finding the initial condition corresponding to a given data sample.

In DEQ Zoo, we refactor DEQ-DDIM~\cite{deq-ddim} through TorchDEQ. 
Given a multi-variate diffusion equilibrium system of $T=1000$ tensors, \cref{Tab:deq-ddim} shows that TorchDEQ accelerates the inversion speed by over $2\times$ than previously appeared under the same Anderson solver configuration while reducing the reconstruction errors, illustrating the utility of TorchDEQ.

\section{Related Works}

\tbf{DEQs}. DEQs have an emerging community. DEQs are receiving growing attention theory-wise~\cite{MON,kawaguchi2020theory,revay2020lipschitz,El2021IDL,feng2021ntk,xie2022optimization}.
Stability~\cite{DEQ_JR,deq-flow} and acceleration~\cite{ham,SamyFPN,PhantomGrad,ramzi2022shine,pal2022mixing,hypersolver} are active research topics in the DEQ community.
DEQs show appealing generalization performance, interpretability~\cite{path-independence}, and robustness~\cite{yang2022adv,yang2023robustness} over semantic segmentation~\cite{MDEQ}, optical flow~\cite{deq-flow}, detection~\cite{wang2020iFPN}, inverse problem~\cite{gilton2021inverse,liu2022online}, meta learning~\cite{swami2021joint}, object-centric learning~\cite{chang2022object}, set prediction~\cite{zhang2022set}, control~\cite{junnarkar2022synthesis}, spiking neural networks~\cite{xiao2021snn}, machine translation~\cite{zheng2023nar}, normalizing flow~\cite{lu2021implicit}, and graph learning~\cite{IGNN,liu2022mgnni}.

\tbf{Libraries}. Scientific software is a driven power that assists deep learning to grow more complex, modular, and large-scale. From fundamental deep learning libraries like \texttt{PyTorch}~\cite{pythorch}, \texttt{Tensorflow}~\cite{tensorflow}, and \texttt{JAX}~\cite{jax2018github} to comprehensive model zoos like \texttt{huggingface}~\cite{huggingface}, and domain-specific libraries like \texttt{fairseq}~\cite{fairseq} for language models, and \texttt{timm}~\cite{timm} for vision backbones, contributions of open-source software are significantly reckoned. 

Recently, there have been many deep learning libraries for neural dynamics like differentiable optimization or differential equations, \eg \texttt{theseus}~\cite{theseus}, \texttt{torchopt}~\cite{torchopt}, \texttt{jaxopt}~\cite{jaxopt}, \texttt{torchdiffeq}~\cite{torchdiffeq}, \texttt{torchdyn}~\cite{torchdyn}, \texttt{betty}~\cite{choe2022betty}, and \texttt{pypose}~\cite{pypose}. 
But none of them is particularly designed for DEQs and is verified to scale up to modern DEQs with good stability, not to mention hosting a model zoo for implicit models.
TorchDEQ and DEQ Zoo step toward this and widely support \sota deep equilibrium models through open-source software.
\vspace{-0.2cm}
\section{Conclusion}
This paper provides a retrospective into Deep Equilibrium Models and identifies that the skillful implementation required can potentially hinder the broader adoption of DEQ.
To remove the invisible barriers and facilitate the utilization of DEQ, we present TorchDEQ, an out-of-the-box PyTorch-based library, and the DEQ Zoo over TorchDEQ, which systematically improves the performance, training stability, and efficiency of DEQ. 
Through these efforts, we aim to foster a thriving DEQ community and make DEQ a more accessible and widely used tool in deep learning.

\bibliography{deq}
\bibliographystyle{plainnat}


\appendix
\clearpage

\section{License}
We have carefully considered the licensing and copyright issues surrounding the implementation of the six published models in DEQ Zoo. We can confirm that the inclusion of these models is compliant with their respective licenses, or we have obtained explicit permission to relicense needed portions of the code under the MIT license as needed. Thus TorchDEQ itself can be released under the MIT license. 

\section{Additional Efficiency Analysis}

TorchDEQ demonstrates superior efficiency or, at the very least, matches the performance of previous implementations. We have specifically focused on optimizing the efficiency of our solvers for both multi-scale and multi-variate equilibrium systems. These systems have historically posed challenges in terms of fixed point solving and differentiation. Consequently, we underscore the efficiency gains achieved over MDEQ~\cite{MDEQ} (multi-scale equilibrium system using Broyden's solver) and DEQ-DDIM~\cite{deq-ddim} (a multi-variate equilibrium system comprising 1000 variables and employing Anderson's solver).
In addition to the efficiency results of DEQ~\cite{DEQ}, MDEQ~\cite{MDEQ}, and DEQ-DDIM~\cite{deq-ddim}, we show further efficiency analysis for the remaining models.

\setlength\tabcolsep{4pt}
\begin{table}[h]
\centering
\caption{\tbf{Efficiency Analysis.} Here, we report both the relative time change and performance gain.}
\vspace{0.1cm}
\begin{tabular}{@{}lcc@{}}
\toprule
Method                               & Relative Time Change  & Relative Performance Gain          \\
\midrule
IGNN (PPI)~\cite{IGNN}               & $+60\%$    & $67\%$   \\
DEQ-Flow-H (KITTI)~\cite{deq-flow}   & $-55\%$    & $14\%$   \\
DEQ-INR (Cat) ~\cite{impsq}          & $-32\%$    & $14\%$   \\
\bottomrule
\end{tabular}
\label{Tab:efficiency}
\end{table}

All models, except for IGNN~\cite{IGNN}, have outperformed their prior counterparts in terms of efficiency and performance by incorporating various techniques available in the library. For IGNN, we introduced new training techniques (Spectral Norm w/ clipping) to enhance stability during the training phase. It is worth noting that our modification led to a $4\%$ absolute performance gain and over a $60\%$ relative error reduction in terms of median Macro F1(\%) across six random seeds, underlining the necessity of our method for improved stability and performance. 
The efficiency for a single run could be improved by excluding the additional technique. Nonetheless, a straightforward measurement for efficiency can be \textit{the time required to achieve stable performance}, which is reduced from six runs to any single run, \ie $160\%\times\frac{1}{6}\approx27\%$.
This reinforces the point that TorchDEQ is not merely an efficient implementation for DEQ models but provides essential methods and insights to ensure stability and performance.

\section{Library Extension}
TorchDEQ offers a modular design with an array of interfaces for library extension. As an example, should a user wish to incorporate a novel computational graph for training, they have the option to inherit from the \texttt{DEQBase} class. Subsequently, their DEQ class can be registered via the \texttt{register\_deq} API, enabling seamless integration with the remaining components of TorchDEQ.

Additionally, the library provides \texttt{register\_solver}, \texttt{register\_norm}, \etc. These APIs afford users the flexibility to add their own fixed-point solvers, normalization, and regularization techniques. Once registered, these user-defined methods can be invoked by other components within the library.

\section{Discussion}
We notice that the importance of the smooth equilibrium landscape~\cite{DEQ_JR,deq-flow,swami2021joint} is usually underestimated in previous literature. Despite mainly working on the library aspect, we hope to bring its importance to the community.

Take DEQ-Flow as an example, as it is trained to test OOD generalization for large-scale real-world challenges. When we train and infer DEQ-Flow-B using fixed point iteration, removing weight normalization can lead to a relative performance drop of over $10\%$ on KITTI~\cite{kitti}, because fixed point iteration cannot solve for the hard equilibrium landscape, and the convergence of DEQ significantly deteriorates. When adopting advanced solvers like the Broyden solver, removing weight normalization leads to no harm. It even slightly improves over the Sintel Clean split~\cite{sintel} (a simpler test set compared to the Sintel Final split and KITTI), as normalization still imposes constraints over model capacity. Advanced solvers can handle the difficulties in the equilibrium landscape, but at the cost of longer training time consumed.

If we scale up to DEQ-Flow-H, a model twice wider than DEQ-Flow-B, things become different. For a high-dimensional fixed point system, no matter using the fixed point iteration or an advanced solver, WN always improves the convergence and thus improves OOD performance, as the equilibrium landscape is easier to solve under normalization. Normalization's advantage outperforms its downside. Regularization techniques like Jacobian regularization and fixed point correction improve DEQs due to a similar reason.

To summarize, if the equilibrium landscape is smooth enough and easy to solve, usually by normalization and regularization, fixed point iteration suffices to train and infer DEQ over high-dimensional systems with a speed bonus, as advanced solvers introduce the cost of matrix multiplication and (approximate) inverse.

However, for some highly structured equilibrium systems, like DEQ-DDIM, inversion using Anderson solvers is still $10\times$ faster than fixed point iteration. This is because the underlying equilibrium landscape of DEQ-DDIM is formed by the training of diffusion models instead of DEQ's training and thus without utilizing landscape regularization. In such a case, advanced solvers like Newton or quasi-Newton methods offer faster convergence rates and practical accelerations through better conditioning on the fixed point system. 

\section{Details for Our DEQ-Transformer}

\setlength\tabcolsep{4pt}
\begin{table}[t]
\centering
\caption{\tbf{Comparison of DEQ Transformer and explicit models.} We report the test Perplexity (PPL) on WikiText-103 and the training iterations (Iters).}
\vspace{0.1in}
\begin{tabular}{@{}lcccc@{}}
\toprule
Method                     & Params &  PPL   &  Iters  \\
\midrule    
Transformer-XL~\cite{trans-xl}       &  151M  & 24.0  & 200K    \\
TaLK Conv~\cite{TaLK}                &  240M  & 23.3  &  -      \\
Transformer-N~\cite{trans-n}         &  148M  & 24.1  & 200K    \\
RFA-GATE~\cite{RFA}                  &  242M  & 23.5  & 150K    \\
Transformer-XL + URPE~\cite{urpe}    &  151M  & 23.2  & 200K    \\
DEQ Transformer~(Ours)               &  98M   & 22.4  & 150K    \\
\bottomrule
\end{tabular}
\vspace{-0.2cm}
\label{Tab:wiki}
\end{table}

For our DEQ transformer, we redesign the model to improve its efficiency, performance, and training speed as transformers gradually become a dominant modeling paradigm in deep learning. Note that we only change the architecture of the DEQ transformer while keeping all other implicit models that appeared in this paper untouched.

We first compare IFT and PG on language modeling using the original training schedule of 300K. We find that IFT's performance will drop when we turn to a shorter schedule like 150K. Then we notice that, if we properly normalize this model, \ie \textit{having a smooth equilibrium landscape}, DEQ transformer can be solved by fixed point iteration while retaining its performance. This saves the training time by a considerable margin as advanced solves naturally introduce overheads~\cite{DEQ_JR}. 

Next, we try to directly differentiate through fixed point iteration, leading to a memory-expensive backpropagation but even more stable training. This is not common in previous DEQs, but we find it helpful for DEQ transformers (and only for DEQ transformers. For example, a 12-step unrolled MDEQ-Tiny can have an accuracy drop by over $1.5\%$ compared to IFT or PG, while a 5-step unrolled MDEQ-Large has a performance drop of over $3\%$ on ImageNet~\cite{MDEQ}). Our goal is to support all the possibilities of DEQs in TorchDEQ. So we particularly optimize the memory usage of unrolled backpropagation in case future DEQs may use this method.

We test different memory managements like gradient checkpoint~\cite{gradckpt} and invertibility~\cite{MacDuvAda2015hyper,sander2021momentum}. It turns out that gradient checkpoint helps reduce the memory usage to a negligible level compared to the original unrolled differentiation at the cost of $1.5\times$ training time. Given the fact that we have optimized the training time by a large margin, this cost is acceptable.
Intriguingly, we find that invertibility is not feasible for DEQs, as a path-independent model can converge the steady fixed point regardless of the initialization, which means DEQ is initialization agnostic and is essentially \textit{not reversible} for the fixed point solving process.
Based on this test, we thus support an interface for gradient checkpointing of DEQs in TorchDEQ.

Having a fast and memory-efficient baseline, we can further optimize the architecture design of our DEQ transformer. We compare the Post Layer Norm~\cite{transformer} and Pre Layer Norm~\cite{xiong2020layer} and find that Pre Layer Norm (LN) can stabilize the DEQ transformer under a higher gradient step. We further add Post Attention LN~\cite{normformer} that stabilizes the training while reducing the perplexity in our test. Additional FFN LN~\cite{normformer,subnorm} can slightly improve the performance but at the cost of $10\%$ more training time (although it adds a small FLOPs number only) and is incompatible with further techniques. So we do not incorporate it into our design.
Other popular transformer techniques, however, cannot show a visible improvement on DEQ transformers, including GELU~\cite{gelu}, Square-ReLU~\cite{sqrelu}, Star-ReLU~\cite{star-relu}, and Global Response Norm~\cite{grn-convnext}, also indicating that designing transformer components universal to different settings can be very challenging.

In the next step, we rearrange the model capacity of DEQ transformers under the same parameters budget.
We notice a 3-block transformer $\f$ design best balances the performance and speed, while a 12-block transformer $\f$ design slightly reduces the perplexity over the 3-block design but drops over $2\times$ slower. Finally, we fix a secret initialization bug that further improves our DEQ transformer.

Altogether, our efforts improve the DEQ transformer to a much stronger performance while retaining satisfactory stability and convergence, requiring only $\nicefrac{1}{3}$ training time of the original DEQ transformer and using even less memory.
As shown in \cref{Tab:wiki}, our DEQ transformer outperforms explicit models by a large margin on long-term dependency modeling.
It is worth noting that explicit transformers employ a context length of 480 or higher for inference on the test set. In contrast, our DEQ transformer has already outperformed the explicit counterparts using a shorter context length of 150 (the attention complexity is dependent on the context length). The test set perplexity can be further reduced to below $\mathbf{22.0}$ when using a context length of 480 for evaluation.

\section{Experiment Settings}
Specifically, we train DEQ transformers only once under different settings because we notice that the results are stable (differences of PPL usually less than $0.2$).

We run over six seeds for MDEQ-Tiny, three for MDEQ-Large on CIFAR-10~\cite{cifar}, and one for MDEQ on ImageNet~\cite{ILSVRC15} (variation of Top-1 accuracy usually less than $0.3$). 

We train IGNN using six random seeds. We apply our clipped DEQ SN to further stabilize the training and enhance its well-posedness, combined with the projected gradient descent from the original paper. Note that the vanilla SN w/o clipping cannot improve its stability.

We train DEQ-Flow-B three times, DEQ-Flow-H once, and DEQ-Flow-H (3$\times$Iters) once. We select the best checkpoint of DEQ-Flow-H (3$\times$Iters) according to the lowest AEPE over the Sintel Final split~\cite{sintel}, making a slightly higher F1-all error on KITTI~\cite{kitti} compared to the default DEQ-Flow-H, although its best F1-all metric can reach $12.5$.
In practice, we find sparse fixed corrections usually suffice, as more correction losses show a ``diminishing marginal return'', \ie as imposing more computing and correction losses, the performance gain gradually diminishes. (Because the convergence gain diminishes, which is a very DEQ-style feature.)
Please note that DEQ-Flow-H (3$\times$Iters) is trained by 360K iterations over the FlyThings3D dataset~\cite{things}, still less than half of the 800K iterations pretraining schedule of transformer-based optical flow models~\cite{gmflow+}.

We follow the official training scripts for DEQ-INR, except for the video reconstruction. We observe a visible inefficiency and slowness of training on the video dataset because reconstructing a high-resolution video using a limited-capacity neural representation can be very challenging.
To ease the training, we apply fixed point correction to video reconstruction using a command of \texttt{---n\_losses 2 ---grad 3}, showing a significantly faster training speed and much better results. (See \cref{Tab:deq-inr}).

We follow the inversion settings, gradient settings, and Anderson mixing configuration of the official DEQ-DDIM code, testing the vanilla DEQ-DDIM and our DEQ-DDIM over the same 100 random seeds. It verifies that the improvements are solely from our library. For this highly structured multi-variate equilibrium model (containing $T=1000$ variables, $\Zstar \in \Rdim{B \times 1000 \times 32 \times 32 \times 3}$), TorchDEQ improves the overall efficiency and performance by over 2$\times$. 
We also try to solve DEQ-DDIM using fixed point iteration to further accelerate it. Unfortunately, this system has a challenging equilibrium landscape and thus cannot be easily solved by fixed point iteration, indicated by the over 10$\times$ inversion time consumed.

\end{document}